\documentclass[lettersize,journal]{IEEEtran}
\usepackage{amsmath,amsfonts}
\usepackage{algorithmic}
\usepackage{array}
\usepackage[caption=false,font=normalsize,labelfont=sf,textfont=sf]{subfig}
\usepackage{textcomp}
\usepackage{stfloats}
\usepackage{url}
\usepackage{verbatim}
\usepackage{graphicx}
\usepackage{cite}
\usepackage{caption}
\usepackage{tabularx}
\usepackage{multirow}
\usepackage{array}
\usepackage{xcolor}
\usepackage{comment}
\usepackage{tabu}
\usepackage{booktabs}
\usepackage{dsfont}% equation symbols
\usepackage{adjustbox}
\usepackage{ragged2e}

\usepackage[lined,boxed,ruled,commentsnumbered]{algorithm2e}
\usepackage{hyperref}
\hypersetup{
    colorlinks=true,
    linkcolor=blue,
    filecolor=magenta,      
    urlcolor=cyan,
    pdftitle={Overleaf Example},
    pdfpagemode=FullScreen,
    }

\newcommand{\ie}{{i.e.}}
\newcommand{\eg}{{e.g.}}
\newcommand{\etc}{{etc}}
\newcommand{\etal}{{et al.}}
\newcommand{\vs}{{vs.}}
\newcommand{\Fig}[1]{Figure \ref{fig:#1}}

\newcommand{\Tab}[1]{Table \ref{tab:#1}}
\newcommand{\Alg}[1]{Algorithm \ref{alg:#1}}

\newcommand{\lab}{l}
\newcommand{\ulab}{u}
\newcommand{\datasets}{\mathcal{X}}

\newcommand{\NW}{\mathcal{W}}
\newcommand{\St}{\mathcal{S}}

\newcommand{\plab}{\hat{l}}

\newcommand{\XL}{\datasets^{\lab}}

\newcommand{\XU}{\datasets^{\ulab}}

\newcommand{\XP}{\datasets^{\plab}}

\newcommand{\HP}{\mathcal{H}}

\newcommand{\HPOD}{\HP_{\NW}}
\newcommand{\HPCoT}{\HP_{ct}}
\newcommand{\HPST}{\HP_{st}}

\newcommand{\Select}[2]{\mbox{{\small Select}}({#1},{#2})}
\newcommand{\Comb}[2]{\mbox{{\small Combination}}({#1},{#2})}

\newcommand{\BaseLineTD}[3]{\mbox{{\small BasicModelTraining}}({#1},{#2},{#3})}
\newcommand{\SelfTraining}[1]{\mbox{{\small SelfTraining}}({#1})}
\newcommand{\TrainD}{\mbox{{\small Train}()}}
\newcommand{\TD}[6]{\mbox{{\small Train}}({#1},{#2},{#3},{#4},{#5},{#6})}
\newcommand{\LastTD}[7]{\mbox{{\small LastTrain}}({#1},{#2},{#3},{#4},{#5},{#6},{#7})}

\newcommand{\RunD}{\mbox{{\small Run}()}}
\newcommand{\RD}[3]{\mbox{{\small Run}}({#1},{#2},{#3})}
\newcommand{\SR}[1]{\mbox{{\small ClassSort}}({#1})}
\newcommand{\Stats}[1]{\mbox{{\small ClassImageList}}({#1})}
\newcommand{\Add}[1]{\mbox{{\small Append}}({#1})}
\newcommand{\Fuse}[2]{\mbox{{\small Fuse}}({#1},{#2})}
\newcommand{\Rand}[2]{\mbox{{\small Rnd}}({#1},{#2})}

\newcommand{\TST}{\mathcal{T}}
\newcommand{\DomFusion}{\mathcal{M}_{df}}
\newcommand{\ClassConfThrs}{\mathcal{V}_{C_T}}
\newcommand{\ClassConfVMap}{\mathcal{V}}

\begin{document}

\title{Co-Training for Unsupervised Domain Adaptation of Semantic Segmentation Models}

\author{Jose L. G\'{o}mez$^{1,2}$,
        Gabriel Villalonga$^{1}$,
        and Antonio M. L\'opez$^{1,2}$
\thanks{$(^{1})$ The authors are  with the Computer Vision Center (CVC) at Universitat Aut\`onoma de Barcelona (UAB), 08193 Bellaterra (Barcelona), Spain.}     \thanks{$(^{2})$ Jose L. Gómez and Antonio M. López are also with the Dpt. of Computer Science at UAB.}
\thanks{The authors acknowledge the support received for this research from the Spanish Grant Ref. PID2020-115734RB-C21 funded by MCIN/AEI/10.13039/501100011033.}
\thanks{Jose L. Gómez acknowledges the financial support to perform his Ph.D. given by the grant FPU16/04131.}
\thanks{Antonio acknowledges the financial support to his general research activities given by ICREA under the ICREA Academia Program.}
\thanks{The authors acknowledge the support of the Generalitat de Catalunya CERCA Program and its ACCIO agency to CVC’s general activities.}
}



\maketitle

\begin{abstract}
Semantic image segmentation is a central and challenging task in autonomous driving, addressed by training deep models. Since this training draws to a curse of human-based image labeling, using synthetic images with automatically generated labels together with unlabeled real-world images is a promising alternative. This implies to address an unsupervised domain adaptation (UDA) problem. In this paper, we propose a new co-training procedure for synth-to-real UDA of semantic segmentation models. It consists of a self-training stage, which provides two domain-adapted models, and a model collaboration loop for the mutual improvement of these two models. These models are then used to provide the final semantic segmentation labels (pseudo-labels) for the real-world images. The overall procedure treats the deep models as black boxes and drives their collaboration at the level of pseudo-labeled target images, {\ie}, neither modifying loss functions is required, nor explicit feature alignment. We test our proposal on standard synthetic and real-world datasets for on-board semantic segmentation. Our procedure shows improvements ranging from {$\sim$13} to {$\sim$31} mIoU points over baselines.
\end{abstract}


\section{Introduction}
Semantic image segmentation is a central and challenging task in autonomous driving, as it involves predicting a class label ( , road, pedestrian, vehicle, {\etc}) per pixel in outdoor images. Therefore, non surprisingly, the development of deep models for semantic segmentation has received a great deal of interest since deep learning is the core for solving computer vision tasks \cite{Long:2015,Badrinarayanan:2017,Chen:2017,Chen:2018,Chen:2018b,Wang:2020,xie:2021}. In this paper, we do not aim at proposing a new deep model architecture for on-board semantic segmentation, but our focus is on the training process of semantic segmentation models. More specifically, we explore the setting where such models must perform in real-world images, while for training them we have access to automatically generated synthetic images with semantic labels together with unlabeled real-world images. It is well-known that training deep models on synthetic images for performing on real-world ones requires domain adaptation \cite{Csurka:2017,Wang:2018}, which must be unsupervised if we have no labels from real-world images \cite{Wilson:2020}. Thus, this paper falls in the realm of unsupervised domain adaptation (UDA) for semantic segmentation \cite{zou:2018,li:2019,Luo:2019,Qin:2019,zou:2019,Yang:2020,Wang:2020a,chao:2021,gao:2021,he:2021,tranheden:2021,zhang:2021}, {\ie}, in contrast to assuming access to labels from the target domain \cite{wang:2020b,chen:2021}. Note that the great relevance of UDA in this context comes from the fact that, until now, pixel-level semantic image segmentation labels are obtained by a cumbersome and error-prone manual work. In fact, this is the reason why the use of synthetic datasets \cite{Ros:2016,Richter:2016,wrenninge:2018} arouses great interest.  

%In summary, we have a supervised source (synthetic) domain, and an unsupervised target (real-world) domain. 

\begin{figure*}[t!]
%[trim=left bottom right top, clip]
\centering
     \includegraphics[width=1.0\textwidth]{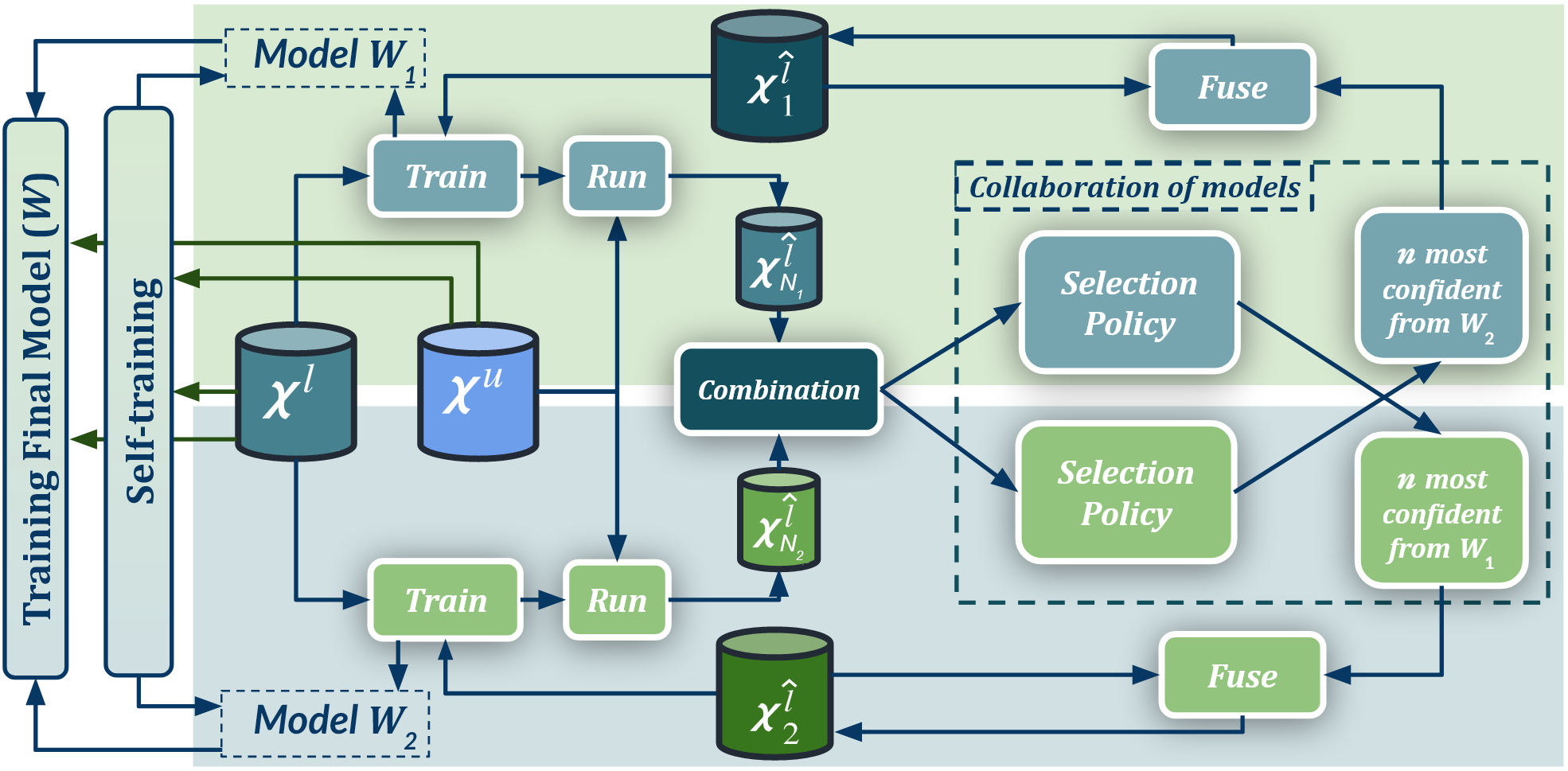}
\caption{Co-training procedure for UDA. $\XL$ is a set of labeled synthetic images, $\XU$ a set of unlabeled real-world images, and $\XP_x$ is the set $x$ of real-world pseudo-labeled images (automatically generated). Our self-training stage provides two initial domain-adapted models ($\NW_1, \NW_2$), which are further trained collaboratively by exchanging pseudo-labeled images. Thus, this procedure treats the deep models as black boxes and drives their collaboration at the level of pseudo-labeled target images, {\ie}, neither modifying loss functions is required, nor explicit feature alignment. See details in Sect. \ref{sec:method} and Algorithms \ref{alg:self-training}--\ref{alg:co-training}.}
\label{fig:cotraining}
\end{figure*}

In this paper, we address synth-to-real UDA following a co-training pattern \cite{Blum:1998}, which is a type of semi-supervised learning (SSL) \cite{Triguero:2015,Engelen:2020} approach. Essentially, canonical co-training consists in training two models in a collaborative manner when only few labeled data are available but we can access to a relatively large amount of unlabeled data. In the canonical co-training paradigm, domain shift between labeled and unlabeled data is not present. However, UDA can be instantiated in this paradigm. 

In previous works, we successfully applied a co-training pattern under the synth-to-real UDA setting for deep object detection \cite{Villalonga:2020,Gomez:2021}. This encourages us to address the challenging problem of semantic segmentation under the same setting by proposing a new co-training procedure, which is summarized in Figure \ref{fig:cotraining}. It consists of a self-training stage, which provides two domain-adapted models, and a model collaboration loop for the mutual improvement of these two models. These models are then used to provide the final semantic segmentation labels (pseudo-labels) for the real-world images. In contrast to previous related works, the overall procedure treats the deep models as black boxes and drives their collaboration only at the level of pseudo-labeled target images, {\ie}, neither modifying loss functions is required, nor explicit feature alignment. We test our proposal on synthetic (GTAV \cite{, Richter:2016}, Synscapes \cite{wrenninge:2018}, SYNTHIA \cite{Ros:2016}) and real-world datasets (Cityscapes \cite{Cordts:2016}, BDD100K \cite{Yu:2020}, Mapillary Vistas \cite{Neuhold:2017}) which have become standard for researching on on-board semantic segmentation. Our procedure shows improvements ranging from {$\sim$13} to {$\sim$31} mean intersection-over-union (mIoU) points over baselines, being less than 10 mIoU points below upper-bounds. Moreover, up to the best of our knowledge, we are the first reporting synth-to-real UDA results for semantic segmentation in BDD100K and Mapillary Vistas.

Section \ref{sec:rw} contextualizes our work. Section \ref{sec:method} details the proposed procedure. Section \ref{sec:er} describes the experimental setup and discusses the obtained results. Finally, Section \ref{sec:conclusions} summarizes this work.

\section{Related works}
\label{sec:rw}
Li {\etal} \cite{li:2019} and Wang {\etal} \cite{Wang:2020a} rely on adversarial alignment to perform UDA. While training a deep model for semantic segmentation, it is performed adversarial image-to-image translation (synth-to-real) together with an adversarial alignment of the model features arising from the \emph{source} (synthetic images) and \emph{target} domains (real-world images). Both steps are alternated as part of an iterative training process. For feature alignment, pseudo-labeling of the target domain images is performed. This involves to apply an automatically computed per-class max probability threshold (MPT) to class predictions. Tranheden {\etal} \cite{tranheden:2021} follow the idea of mixing source and target information (synthetic and real) as training samples \cite{Ros:2016}. However, target images are used after applying ClassMix \cite{Olsson:2021}, {\ie}, a class-based collage between source and target images. This requires the semantic segmentation ground truth, which for the synthetic images (source) is available while for the real-world ones (target) pseudo-labels are used. Such domain adaptation via cross-domain mixed sampling (DACS) is iterated so that the semantic segmentation model can improve its accuracy by eventually producing better pseudo-labels. Gao {\etal} \cite{gao:2021} not only augment target images with source classes, but the way around too. Their dual soft-paste (DSP) is used withing a teacher-student framework, where the teacher model generates the pseudo-labels. Zou {\etal} \cite{zou:2018} propose a self-training procedure where per-cycle pseudo-labels are considered by following a self-paced curriculum learning policy. An important step is class-balanced self-training (CBST), which is similar to MPT since a per-class confidence-based selection of pseudo-labels is performed. Spatial priors (SP) based on the source domain (synth) are also used. The authors improved their proposal in \cite{zou:2019} by incorporating confidence regularization steps for avoiding error drift in the pseudo-labels. 

Chao {\etal} \cite{chao:2021} assume the existence of a set of semantic segmentation models independently pre-trained according to some UDA technique. Then, the pseudo-label confidences coming from such models are unified, fused, and finally distilled into a student model. Zhang {\etal} \cite{zhang:2021} propose a multiple fusion adaptation (MFA) procedure, which integrates online-offline masked pseudo-label fusion, single-model temporal fusion, and cross-model fusion. To obtain the offline pseudo-labels, existing UDA methods must be applied. In particular, so-called FDA \cite{Yang:2020} method is used to train two different models which produce two maps of offline pseudo-labels for each target image. Other two models, $m_1 \& m_2$, are then iteratively trained. Corresponding temporal moving average models, $\hat{m}_1 \& \hat{m}_2$, are kept and used to generate the online pseudo-labels. The training total loss seeks for consistence between class predictions of each $m_i$ and both offline pseudo-labels and class predictions from the corresponding $\hat{m}_i$. Moreover, consistence between the online pseudo-labels from $\hat{m}_i$ and the predictions from $m_j$, $i \neq j$, is used as a collaboration mechanism between models. Offline and online pseudo-labels are separately masked out by corresponding CBST-inspired procedures. He {\etal} \cite{he:2021} assume the existence of different source domains. To reduce the visual gap between each source domain and the target domain there is a first step where their LAB color spaces are aligned. Then, there are as many semantic segmentation models to train as source domains. Model training relies on source labels and target pseudo-labels. The latter are obtained by applying the model to the target domain images and using a CBST-inspired procedure for thresholding the resulting class confidences. The training of each model is done iteratively so that the relevance of pseudo-labels follows a self-paced curriculum learning. Collaboration between models is also part of the training. In particular, it is encouraged the agreement of the confidences of the different models when applied to the same source domain, for all source domains. Qin {\etal} \cite{Qin:2019} proposed a procedure consisting of feature alignment based on cycleGAN \cite{Zhu:2017}, additional domain alignment via two models whose confidence discrepancies are considered, and a final stage where the confidences of these models are combined to obtain pseudo-labels which are later used to fine-tune the models. Luo {\etal} \cite{Luo:2019} focused on the lack of semantic consistency (some classes may not be well aligned between source and target domains, while others can be). Rather than a global adversarial alignment between domains, a per-class adversarial alignment is proposed. Using a common feature extractor, but two classification heads, per-class confidence discrepancies between the heads are used to evaluate class alignment. The classification heads are forced to be different by a cosine distance loss. Combining the confidences of the two classifiers yields the final semantic segmentation prediction. This approach does not benefit from pseudo-labels.

In contrast to these methods, our proposal is purely data-driven in the sense of neither requiring changing the loss function of the selected semantic segmentation model, nor explicit model features alignment of source and target domains via loss function, {\ie}, we treat the semantic segmentation model as a black box. Our UDA is inspired in co-training \cite{Blum:1998}, so we share with some of the reviewed works the benefit of leveraging pseudo-labels. In our proposal two models collaborate at pseudo-label level for compensating labeling errors. These two models arise from our previous self-training stage, which share with previous literature self-paced adaptive thresholding inspired by MPT and CBST, as well as pixel-level domain mixes inspired by ClassMix. Our proposal is complementary to image pre-processing techniques such as color space adjustments and learnable image-to-image transformations. In case of having multiple synthetic domains, we assume they are treated as a single (heterogeneous) source domain, which has been effective in other visual tasks \cite{Ranftl:2022}. 

\section{Method}
\label{sec:method}

In this section, we explain our data-driven co-training procedure, {\ie}, the self-training stage, and the model collaboration loop for the mutual improvement of these two models, which we call \emph{co-training loop}. Overall, our proposal works at pseudo-labeling level, {\ie}, it does not change the loss function of the semantic segmentation model under training. Global transformations ({\eg}, color corrections, learnable image-to-image transformations) on either source or target domain images are seen as pre-processing steps. Moreover, in case of having access to multiple synthetic datasets, whether to use them one at a time or simultaneously is just a matter of the input parameters passed to our co-training procedure. 

\begin{algorithm}[t!]
\caption{Self-training Stage}
\label{alg:self-training}
\DontPrintSemicolon
\SetKwInOut{Input}{Input}\SetKwInOut{Output}{Output}
\Input{{\small {Set of labeled images:} $\XL$\\ 
       {Set of unlabeled images:} $\XU$\\
       {Net. init. weights \& training hyp.-p.:} $\NW, \HPOD$\\
       {Self-t. hyp.-p.:} $\HPST=\{\TST,N,n,K_m,K_M,\DomFusion\}$\\}} 
\Output{{\small {Two refined models:} ${\NW}_{K_m}, {\NW}_{K_M}$}}
{\small
\begin{tabularx}{\textwidth}{rl}
\multicolumn{2}{c}{// Initialization}\\
$\NW_0$                      & $\gets$ $\BaseLineTD{\NW}{\HPOD}{\XL}$\\
$<\XP,k,\ClassConfThrs,\NW>$ & $\gets$ $<\emptyset,0,{\bf 0},\NW_0>$\\
\end{tabularx}}
\Repeat{{\small $({k == {K_M}} ; {k\mbox{++}})$}}
{
{\small
\begin{tabularx}{\textwidth}{rl}
\multicolumn{2}{c}{// Self-training Loop} \\
$<\XP_{N},\ClassConfThrs>$   & $\gets$ $\RD{\NW}{\Rand{\XU}{N}{}}{k,\TST}$\\
$\XP$                        & $\gets$ $\Fuse{\XP}{\Select{n}{\XP_{N}}}$\\
$\NW$                        & $\gets$ $\TD{{\NW}_0}{\HPOD}{\XL}{\XP}{\DomFusion}{\ClassConfThrs}$\\
\end{tabularx}
{\bf if} $(k=={K_m})$ {\bf then} ${\NW}_{K_m} \gets \NW$\\ 
{\bf elseif} $(k=={K_M})$ {\bf then} ${\NW}_{K_M} \gets \NW$ {\bf endif}
}}
\KwRet{{\small ${\NW}_{K_m}, {\NW}_{K_M}$}}
\end{algorithm}

\subsection{Self-training stage}

\Alg{self-training} summarizes the self-training stage, which we detail in the following.\\

\noindent\textbf{Input \& output parameters.} The input  refers to the set of fully labeled source images; while $\XU$ refers to the set of unlabeled target images. In our UDA setting, the source images are synthetic and have automatically generated per-pixel semantic segmentation ground truth (labels), while the target images are acquired with real-world cameras. $\NW$ refers to the weights of the semantic segmentation model (a CNN) already initialized (randomly or by pre-training on a previous task); while $\HPOD$ are the usual hyper-parameters required for training the model in a supervised manner ({\eg}, optimization policy, number of iterations, {\etc}). $\HPST=\{\TST,N,n,K_m,K_M,\DomFusion\}$ consists of parameters specifically required by the proposed self-training. $K_M$ is the number of self-training cycles, where we output the model, ${\NW}_{K_M}$, at the final cycle. $K_m$, $K_m<K_M$, indicates an intermediate cycle from where we also output the corresponding model, ${\NW}_{K_m}$. $N$ is the number of target images used to generate pseudo-labels at each cycle, while $n$, $n<N$, is the number of pseudo-labeled images to be kept for next model re-training. $\HPST$ also contains $\TST=\{p_m,p_M,{\Delta p},C_m,C_M\}$, a set of parameters to implement a self-paced curriculum learning policy for obtaining pseudo-labels from model confidences, which is inspired in MPT \cite{li:2019} and CBST \cite{zou:2018}. Finally, $\DomFusion=\{p_{MB},p_{CM}\}$ consists of parameters to control how source and target images are combined.\\

\noindent\textbf{Initialization.} We start by training a model, $\NW_0$, on the (labeled) source images, $\XL$, according to $\NW$ and $\HPOD$. At each self-training cycle, $\NW_0$ is used as pre-trained model.

\noindent\textbf{Self-training cycles (loop).} Each cycle starts by obtaining a set of pseudo-labeled images, $\XP_{N}$. For the shake of speed, we do not consider all the images in $\XU$ as candidates to obtain pseudo-labels. Instead, $N$ images are \emph{selected} from $\XU$ and, then, the current model $\NW$ is applied to them (\emph{run}). Thus, we obtain $N$ semantic maps. Each map can be seen as a set of confidence channels, one per class. Thus, for each class, we have $N$ confidence maps. Let's term as $\ClassConfVMap_c$ the vector of confidence values $>0$ gathered from the $N$ confidence maps of class $c$. For each class $c$, a confidence threshold, $C_{T_c}$, is set as the value required for having $p\%$ values of vector $\ClassConfVMap_c$ over it, where $p=\min(p_m+k{\Delta p},p_M)$. Let's term as $\ClassConfThrs$ the vector of confidence thresholds from all classes. Now, $\ClassConfThrs$ is used to perform per-class thresholding on the $N$ semantic segmentation maps, so obtaining the $N$ pseudo-labeled images forming $\XP_{N}$. Note how the use of $p_m+k{\Delta p}$, where $k$ is the self-training cycle, acts as a mechanism of \emph{self-paced curriculum learning} on the thresholding process. The maximum percentage, $p_M$, allows to prevent noise due to accepting too much per-class pseudo-labels eventually with low confidence.  Moreover, for any class $c$, we apply the rule $C_{T_c}\gets\max(C_m,\min(C_{T_c},C_M))$; where, irrespective of $p\%$, $C_m$ prevents from considering not sufficiently confident pseudo-labels, while $C_M$ ensures to consider pseudo-labels with a sufficiently high confidence. Then, in order to set the final set of pseudo-labels during each cycle, only $n$ of the $N$ pseudo-labeled images are \emph{selected}. In this case, an image-level confidence ranking is established by simply averaging the confidences associated to the pseudo-labels of each image. The {top-$n$} most confident images are considered and \emph{fused} with images labeled in previous cycles. If one of the selected $n$ images was already incorporated in previous cycles, we kept the pseudo-labels corresponding to the highest image-level confidence average. The resulting set of pseudo-labeled images is termed as $\XP$.

Finally, we use the (labeled) source images, $\XL$, and the pseudo-labeled target images, $\XP$, to \emph{train} a new model, $\NW$, by fine-tuning $\NW_0$ according to the hyper-parameters $\HPOD$ and $\DomFusion$. A parameter we can find in any $\HPOD$ is the number of images per mini-batch, $N_{MB}$. Then, given $\DomFusion=\{p_{MB},p_{CM}\}$, for training $\NW$ we use $p_{MB}N_{MB}$ images from $\XP$ and the rest from $\XL$. In fact, the former undergo a ClassMix-inspired collage transform \cite{Olsson:2021}. In particular, we select $p_{MB}N_{MB}$ images from $\XL$ and, for each one individually, we gather all the class information (appearance and labels) until considering a $p_{CM}\%$ of classes, going from less to more confident ones, which is possible thanks to $\ClassConfThrs$. This information is pasted at appearance level (class regions from source on top of target images) and at label level (class labels from source on top of the pseudo-label maps of the target images).

\begin{algorithm}[t!]
\caption{Collaboration of Models}
\label{alg:collaboration}
\DontPrintSemicolon
\SetKwInOut{Input}{Input}\SetKwInOut{Output}{Output}
\Input{{\small {Sets of pseudo-labeled images:} $\XP_{N_1}, \XP_{N_2}$\\ 
       {Vectors of per-class conf. thr.:} ${\ClassConfThrs}_1,{\ClassConfThrs}_2$\\
       {Amount of images to exchange:} $n$\\
        {Image-level confidence threshold control:} $\lambda$\\}}
\Output{{\small {New sets of ps.-lab. images:} $\XP_{N_1,new}, \XP_{N_2,new}$}}
\scalebox{0.81}[0.81]{\begin{tabularx}{\textwidth}{rl}
\multicolumn{2}{l}{// $\SR{v}$ returns the vector of sorted $v$ indices after}\\
\multicolumn{2}{l}{// sorting by $v$ values, so that $\Delta_i[k]$ is a class index.}\\
$\Delta_1$ & $\gets$ $\SR{{\ClassConfThrs}_2-{\ClassConfThrs}_1}$\\
$\Delta_2$ & $\gets$ $\SR{{\ClassConfThrs}_1-{\ClassConfThrs}_2}$\\
\multicolumn{2}{l}{// $\Stats{\datasets}$ returns a vector so that $\St_i[k]$ is the}\\
\multicolumn{2}{l}{// list of images in $\datasets$ containing pseudo-labels of class $k$.}\\
$\St_1$    & $\gets$ $\Stats{\XP_{N_1}},$  \\
$\St_2$    & $\gets$ $\Stats{\XP_{N_2}}$\\
$\XP_{N_1,new},\XP_{N_2,new}$  & $\gets$ $\emptyset,\emptyset$\\
$k,N_c$    & $\gets$ $0,\mbox{Num. Classes}$\\
\end{tabularx}}
\Repeat{{\small $((|\XP_{N_1,new}|==|\XP_{N_2,new}|==n) \| (k == N_c) ; {k\mbox{++}})$}}
{
\scalebox{0.84}[0.84]{\begin{tabularx}{\textwidth}{rl}
\multicolumn{2}{c}{// $\St_{j}[\Delta_i[k]] > t$ applies element-wise.}\\
\multicolumn{2}{c}{// Images from $\XP_{N_2}$ move to $\XP_{N_1,new}$.}\\
$t_{c}$    & $\gets$ $\lambda\max(\St_{1}[\Delta_2[k]]) + (1-\lambda) \min(\St_{1}[\Delta_2[k]])$ \\
$\XP_{N_1,new}$ & $\gets$ $\Add{\XP_{N_1,new},\St_{1}[\Delta_2[k]] > t_{c}}$\\
\multicolumn{2}{c}{// Images from $\XP_{N_1}$ move to $\XP_{N_2,new}$.}\\
$t_{c}$    &  $\gets$ $\lambda\max(\St_{2}[\Delta_1[k]]) + (1-\lambda) \min(\St_{2}[\Delta_1[k]])$\\
$\XP_{N_2,new}$ & $\gets$ $\Add{\XP_{N_2,new},\St_{2}[\Delta_1[k]] > t_{c}}$\\
\end{tabularx}}
}
\KwRet{{\small $\XP_{N_1,new}, \XP_{N_2,new}$}}
\end{algorithm}

\begin{algorithm}[h]
\caption{Co-training Procedure\\ \textbf{Uses:} Algorithms \ref{alg:self-training} \& \ref{alg:collaboration}.}
\label{alg:co-training}
\DontPrintSemicolon
\SetKwInOut{Input}{Input}\SetKwInOut{Output}{Output}
\Input{{\small {Set of labeled images:} $\XL$\\ 
       {Set of unlabeled images:} $\XU$\\
       {Net. init. weights \& training hyp.-p.:} $\NW, \HPOD$\\
       {Self-t. hyp.-p.:} $\HPST=\{\TST,N,n,K_m,K_M,\DomFusion\}$\\
       {Co-t. hyp.-p.:} $\HPCoT=\{K,w,\lambda\}$\\}
\Output{{\small {Refined model:} ${\NW}$}}}
\scalebox{0.81}[0.81]{\begin{tabularx}{\textwidth}{rl}
\multicolumn{2}{c}{// Initialization}\\
$\NW_{0,1},\NW_{0,2}$                                             & $\gets$ $\SelfTraining{\XL,\XU,\NW,\HPOD,\HPST}$\\
$\XP_1,\XP_2,k,{\ClassConfThrs}_1,{\ClassConfThrs}_2,\NW_1,\NW_2$ & $\gets$ $\emptyset,\emptyset,0,{\bf 0},{\bf 0},\NW_{0,1},\NW_{0,2}$\\
\end{tabularx}}
\Repeat{{\small $({k == K} ; {k\mbox{++}})$}}
{
\scalebox{0.84}[0.84]{\begin{tabularx}{\textwidth}{rl}
\multicolumn{2}{c}{// Co-training Loop} \\
$\XU_N$                           & $\gets$ $\Rand{\XU}{N}$\\
$<\XP_{N_1},{\ClassConfThrs}_1>$  & $\gets$ $\RD{\NW_1}{\XU_N}{k,\TST}$\\
$<\XP_{N_2},{\ClassConfThrs}_2>$  & $\gets$ $\RD{\NW_2}{\XU_N}{k,\TST}$\\
$\XP_{N_1},\XP_{N_2}$             & $\gets$ $\Comb{\XP_{N_1}}{\XP_{N_2}}$\\
$\XP_{N_1},\XP_{N_2}$             & $\gets$ ${\mbox{{\small Collaboration}}}(\XP_{N_1},{\ClassConfThrs}_1,\XP_{N_2},{\ClassConfThrs}_2,{n, \lambda})$\\
$\XP_1, \XP_2$                    & $\gets$ $\Fuse{\XP_1}{\XP_{N_1}}, \Fuse{\XP_2}{\XP_{N_2}}$\\
$\NW_1$                           & $\gets$ ${\mbox{{\small Train}}}(\NW_{0,1},\HPOD,\XL,\XP_1,\DomFusion,{\ClassConfThrs}_1)$\\
$\NW_2$                           & $\gets$ ${\mbox{{\small Train}}}(\NW_{0,2},\HPOD,\XL,\XP_2,\DomFusion,{\ClassConfThrs}_2)$\\
\end{tabularx}}
}
{\small $\NW$ $\gets$ $\LastTD{w}{\NW_1}{\NW_2}{\HPOD}{\XL}{\XU}{\DomFusion}$}\\
\KwRet{{\small $\NW$}}
\end{algorithm}

\subsection{Co-training procedure}

\Alg{co-training} summarizes the co-training procedure supporting the scheme shown in \Fig{cotraining}, which is based on the previous self-training stage (\Alg{self-training}), on combining pseudo-labels, as well as on a model collaboration stage (\Alg{collaboration}). We detail \Alg{co-training} in the following.

\noindent\textbf{Input \& output parameters, and Initialization.} Since the co-training procedure includes the self-training stage, we have the input parameters required for \Alg{self-training}. As additional parameters we have $\HPCoT=\{K,w,\lambda\}$, where $K$ is the maximum number of iterations for mutual model improvement, which we term as \emph{co-training loop}, $w$ is just a selector to be used in the last training step (after the co-training loop), and $\lambda$ is used during pseudo-label exchange between models. The output parameter, ${\NW}$, is the final model. The co-training procedure starts by running the self-training stage.\\

\noindent\textbf{Co-training cycles (loop).} Similarly to self-training, a co-training cycle starts by obtaining pseudo-labeled images. In this case two sets, $\XP_{N_1} \& \XP_{N_2}$, are obtained since we \emph{run} two different models, $\NW_1 \& \NW_2$. These are applied to the same subset, $\XU_N$, of $N$ unlabeled images \emph{randomly selected} from $\XU$. As for self-training, we not only obtain $\XP_{N_1} \& \XP_{N_2}$, but also corresponding vectors of per-class confidence thresholds, ${\ClassConfThrs}_1 \& {\ClassConfThrs}_2$. Since, $\XP_{N_1} \& \XP_{N_2}$ come from the same $\XU_N$ but result from different models, we can perform a simple step of pseudo-label \emph{combination}. In particular, for each image in $\XP_{N_i}$, if a pixel has the \emph{void} class as pseudo-label, then, if the pseudo-label for the same pixel of the corresponding image in $\XP_{N_j}$ is not \emph{void}, we adopt such pseudo-label, $i\in\{1,2\}, j\in\{1,2\}, i\neq j$. This step reduces the amount of non-labeled pixels, while keeping pseudo-labeling differences between $\XP_{N_1} \& \XP_{N_2}$ at non-void pseudo-labels. 

Note that co-training strategies assume that the models under collaboration perform in a complementary manner. Therefore, after this basic combination of pseudo-labels, a more elaborated \emph{collaboration} stage is applied, which is described in \Alg{collaboration}. Essentially, $n$ pseudo-labeled images from $\XP_{N_i}$ will form the new $\XP_{N_j}$ after such collaboration, $i\in\{1,2\}, j\in\{1,2\}, i\neq j$. Thus, along the co-training cycle, pseudo-labeled images arising from $\NW_i$ will be used to retain $\NW_j$. In particular, visiting first those images containing classes for which $\XP_{N_i}$ is more confident than $\XP_{N_j}$, sufficiently high confident images in $\XP_{N_i}$ are selected for the new $\XP_{N_j}$ set, until reaching $n$. The class confidences of $\XP_{N_i} \& \XP_{N_j}$ are given by the respective ${\ClassConfThrs}_1 \& {\ClassConfThrs}_2$, while the confidence of a pseudo-labeled image is determined as the average of the confidences of its pseudo-labels. Being sufficiently high confident means that the average is over a dynamic threshold controlled by the $\lambda$ parameter. 

Once this process is finished, we have two new sets of pseudo-labels, $\XP_{N_1} \& \XP_{N_2}$, which are used separately for finishing the co-training cycle. In particular, each new $\XP_i$ is used as its self-training counterpart (see $\XP$ in the loop of \Alg{self-training}), {\ie}, performing the fusion with the corresponding set of pseudo-labels from previous cycles and fine-tuning of $\NW_{0,i}$. Finally, once the co-training loops finishes, a last train is performed. In this case, the full $\XU$ is used to produce pseudo-labels. For this task, we can use an ensemble of $\NW_1$ and $\NW_2$ ({\eg}, averaging confidences), or any of these two models individually. This option is selected according to the parameter $w$. In this last training, the ClassMix-inspired procedure is not applied, but mixing source and target images at mini-batch level still is done according to the value $p_{MB}\in\DomFusion$. It is also worth noting that, inside co-training loop, the two $\RunD$ operations can be parallelized, and the two $\TrainD$ too.

\section{Experimental results}
\label{sec:er}

\subsection{Datasets and evaluation}
Our experiments rely on three well-known synthetic datasets used for UDA semantic segmentation as source data, namely, GTAV \cite{Richter:2016}, SYNTHIA \cite{Ros:2016} and Synscapes \cite{wrenninge:2018}. GTAV is composed by 24,904 images with a resolution of $1914\times1052$ pixels directly obtained from the render engine of the videogame GTA V. Synscapes is composed by 25,000 images with a resolution of $1440\times720$ pixels of urban scenes, obtained by using a physic-based rendering pipeline. SYNTHIA is composed by 9,000 images of urban scenes highly populated, with a resolution of $1280\times760$ pixels, generated by a videogame-style rendering pipeline based on the Unity3D framework. As real-world datasets (target domain) we rely on Cityscapes \cite{Cordts:2016}, BDD100K \cite{Yu:2020} and Mapillary Vistas \cite{Neuhold:2017}. Cityscapes is a popular dataset composed of on-board images acquired at different cities in Germany under clean conditions ({\eg}, no heavy occlusions or bad weather), it is common practice to use 2,975 images for training semantic segmentation models, and 500 images for reporting quantitative results. The latter are known as validation set. Cityscapes images have a resolution of $2048\times1024$ pixels. Another dataset is BDD100K, which contains challenging on-board images taken from different vehicles, in different US cities, and under diverse weather conditions. The dataset is divided into 7,000 images for training purposes and 1,000 for validation. However, a high amount of training images are heavily occluded by the ego vehicle, thus, for our experiments we rely on a occlusion-free training subset of 1,777 images. Image resolution is $1280\times720$ pixels. Finally, Mapillary Vistas is composed by high resolution images of street views around the world. These images have a high variation in resolutions and aspect ratios due to the fact that are taken from diverse devices like smartphones, tablets, professional cameras, {\etc}. For simplicity, we only consider those images with aspect ratio 4:3, which, in practice, are more than the 75\%. Then, we have 14,716 images for training and 1,617 for validation. 

As is common practice, we evaluate the performance of our system on the validation set of each real-world (target) dataset using the 19 official classes defined for Cityscapes. These 19 classes are common in all the datasets except in SYNTHIA that only contains 16 of these 19 classes, additional dataset-specific classes are ignored for training and evaluation. Note that, although there are semantic labels available for the target datasets, for performing UDA we ignore them at training time, and we use them at validation time. In other words, we only use the semantic labels of the validation sets, and with the only purpose of reporting quantitative results. All the synthetic datasets provide semantic labels, since they act as source domain, we use them. In addition, we note that for our experiments we do not perform any learnable image-to-image transform to align synthetic and real-world domains (like GAN-based ones). However, following \cite{he:2021}, we perform synth-to-real LAB space alignment as pre-processing step.

As is standard, quantitative evaluation relies on PASCAL VOC intersection-over-union metric $IoU = TP/(TP + FP + FN)$ \cite{everingham:2015}, where TP, FP, and FN refer to true positives, false positives, and false negatives, respectively. IoU can be computed per class, while using a mean IoU (mIoU) to consider all the classes at once.
 
\begin{table*}[t]
%\centering
\caption{Hyper-parameters of our method (Sect. \ref{sec:method} and Alg. \ref{alg:self-training}--\ref{alg:co-training}). Datasets: GTAV (G), Synscapes (S), STNTHIA (SIA), Cityscapes (C), BDD (B), Mapillary Vistas (M).}\label{tab:parameters}
\resizebox{1.0\linewidth}{!}
{ \setlength{\tabcolsep}{1.5mm} 
\begin{tabular}{@{}cc|cccccc|cc|cccc|ccccc@{}}
\toprule
       &     &  &    &   &  &   & & \multicolumn{2}{c|}{\boldmath{$\DomFusion$}}   & \multicolumn{4}{c|}{\boldmath{$\HPST$}}   & \multicolumn{5}{c}{\boldmath{$\HPCoT$}} \\ 
\boldmath{$Source$}  & \boldmath{$Target$}       & \boldmath{$N$}   & \boldmath{$n$} & \boldmath{${\Delta p}$} & \boldmath{$C_m$} & \boldmath{$C_M$} & \boldmath{$N_{MB}$}  & \boldmath{$p_{MB}$} & \boldmath{$p_{CM}$}  & \boldmath{$P_m$}   & \boldmath{$P_M$} & \boldmath{$K_m$} & \boldmath{$K_M$} & \boldmath{$P_m$} & \boldmath{$P_M$} & \boldmath{$K$} & \boldmath{$w$} & \boldmath{$\lambda$} \\
\midrule
SIA     & C     & 500   & 100 & 0.05 & 0.5 & 0.9 & 4 & 0.75 & 0.5  & 0.5        & 0.6            & 1        & 10        & 0.5              & 0.6  &   5    &  1 & 0.8 \\
S, G+S  & C, M  & 500   & 100 & 0.05 & 0.5 & 0.9 & 4 & 0.75 & 0.5  & 0.5        & 0.6            & 1        & 10        & 0.5              & 0.6  &   5    &  1 & 0.8 \\
G       & C     & 500   & 100 & 0.05 & 0.5 & 0.9 & 4 & 0.75 & 0.5  & 0.3        & 0.5            & 1        & 10        & 0.5              & 0.6  &   5    &  1 & 0.8 \\
G+S     & B     & 500   & 100 & 0.05 & 0.5 & 0.9 & 2 & 0.5  & 0.5  & 0.3        & 0.5            & 1        & 10        & 0.5              & 0.6  &   5    &  1 & 0.8 \\
\bottomrule
\end{tabular}
}
\end{table*}

\subsection{Implementation details}

We use the Detectron2 \cite{wu:2019detectron2} framework and leverage their implementation of DeepLabV3+ for semantic segmentation, with ImageNet weight initialization. We chose V3+ version of DeepLab instead the V2 because it provides a configuration which fits well in our 12GB-memory GPUs, turning out in a $\times2$ training speed over the V2 configuration and allowing a higher batch size. Other than this, V3+ does not provide accuracy advantages over V2. We will see it when discussing \Tab{sota_comp}, where the baselines of V3+ and V2 performs similarly (SYNTHIA case) or V3+ may perform worse (GTAV case). The hyper-parameters used by our co-training procedure are set according to \Tab{parameters}. Since their meaning is intuitive, we just tested some reasonable values, but did not perform hyper-parameter search. As we can see in \Tab{parameters} they are pretty similar across datasets. This table does not include the hyper-parameter related to the training of DeepLabV3+, termed as $\HPOD$ in Algorithms \ref{alg:self-training}--\ref{alg:co-training}, since they are not specific of our proposal. Thus, we summarize them in the following. 

For training the semantic segmentation models, we use SGD optimizer with a starting learning rate of 0.002 and momentum 0.9. We crop the training images to $1024\times512$ pixels, $816\times608$, and $1280\times720$, when we work with Cityscapes, Mapillary Vistas, and BDD100K, respectively. Considering this cropping and our available hardware, we set batch sizes ($N_{MB}$) of 4 images, 4, and 2, for these datasets, respectively. Moreover, we perform data augmentation consisting of random zooms and horizontal flips. For computing each source-only baseline model ($\NW_0$ in \Alg{self-training}) and the final model (returned $\NW$ in \Alg{co-training}) we use a two-step learning rate decay of 0.1 at 1/3 and 2/3 of the training iterations. In these cases, the number of iterations are set to 60K when we work with Cityscapes and Mapillary Vistas, and 120K for BDD100K to maintain consistency given the mentioned batch sizes. The number of iterations for the self-training stage and the co-training loop are equally set to 8K for Cityscapes and Mapillary Vistas, and 16K for BDD100K. 

For training only using GTAV, a class balancing sample policy (CB) is applied. Due to the scarcity of samples from several classes ({\eg}, bicycle, train, rider and motorcycle), these are under-represented during training. A simple, yet efficient, method to balance the frequency of samples from these classes is computing individual class frequency in the whole training dataset and apply a higher selection probability for the under-represented classes. The other synthetic datasets in isolation and the combination of GTAV + Synscapes are already well balanced and we do not need to apply this technique.

\begin{table*}[t]
\caption{UDA results. mIoU considers the 19 classes. mIoU* considers 13 classes, which only applies to SYNTHIA, where classes with '*' are not considered for global averaging and those with '-' scores do not have available samples. \textbf{Bold} stands for best, and \underline{underline} for second best. In this table, the target domain is always Cityscapes.}
\label{tab:sota_comp}
\centering
\setlength{\tabcolsep}{1.0mm} 
\scalebox{0.97}[0.97]{\begin{tabular}{l||ccccccccccccccccccc||c||c|c|c||}
Methods & \rotatebox[origin=c]{90}{Road} & \rotatebox[origin=c]{90}{Sidewalk} & \rotatebox[origin=c]{90}{Building} & \rotatebox[origin=c]{90}{Wall*} & \rotatebox[origin=c]{90}{Fence*} & \rotatebox[origin=c]{90}{Pole*} & \rotatebox[origin=c]{90}{Traffic light} & \rotatebox[origin=c]{90}{Traffic Sign} & \rotatebox[origin=c]{90}{Vegetation} & \rotatebox[origin=c]{90}{Terrain} & \rotatebox[origin=c]{90}{Sky} & \rotatebox[origin=c]{90}{Person} & \rotatebox[origin=c]{90}{Rider} & \rotatebox[origin=c]{90}{Car} & \rotatebox[origin=c]{90}{Truck} & \rotatebox[origin=c]{90}{Bus} & \rotatebox[origin=c]{90}{Train} & \rotatebox[origin=c]{90}{Motorbike} & \rotatebox[origin=c]{90}{Bike} & \rotatebox[origin=c]{90}{mIoU*} & \rotatebox[origin=c]{90}{mIoU} & \rotatebox[origin=c]{90}{Baseline} & \rotatebox[origin=c]{90}{$\Delta$(Diff.)} \\
\midrule
\multicolumn{24}{c}{SYNTHIA (Source) $\rightarrow$ Cityscapes} \\
\midrule
AdSegNet\cite{tsai:2018} & 81.7 & 39.1 & 78.4 & 11.1 & 0.3 & 25.8 & 6.8 & 9.0 & 79.1 & - & 80.8 & 54.8 & 21.0 & 66.8 & - & 34.7 & - & 13.8 & 29.9 & 45.8 & 39.6 & 33.5 & +6.1 \\
IntraDA \cite{pan:2020} & 84.3 & 37.7 & 79.5 & 5.3 & 0.4 & 24.9 & 9.2 & 8.4 & 80.0 & - & 84.1 & 57.2 & 23.0 & 78.0 & - & 38.1 & - & 20.3 & 36.5 & 48.9 & 41.7 & 33.5 & +8.2\\
CBST \cite{zou:2018} & 68.0 & 29.9 & 76.3 & 10.8 & 1.4 & 33.9 & 22.8 & 29.5 & 77.6 & - & 78.3 & 60.6 & 28.3 & 81.6 & - & 23.5 & - & 18.8 & 39.8 & 48.9 & 42.6 & 29.2 & +13.4 \\
CRST \cite{zou:2019} & 67.7 & 32.2 & 73.9 & 10.7 & 1.6 & 37.4 & 22.2 & 31.2 & 80.8 & - & 80.5 & 60.8 & 29.1 & 82.8 & - & 25.0 & - & 19.4 & 45.3 & 50.1 & 43.8 & \underline{34.9} & +8.9 \\
DACS \cite{tranheden:2021} & 80.5 & 25.1 & 81.9 & 21.4 & 2.8 & 37.2 & 22.6 & 23.9 & 83.6 & - & 90.7 & 67.6 & 38.3 & 82.9 & - & 38.9 & - & 28.4 & 47.5 & 54.8 & 48.3 & 29.4 & +18.9 \\
DSP \cite{gao:2021} & 86.4 & 42.0 & 82.0 & 2.1 & 1.8 & 34.0 & 31.6 & 33.2 & 87.2 & - & 88.5 & 64.1 & 31.9 & 83.8 & - & 65.4 & - & 28.8 & 54.0 & 59.9 & 51.0 & 33.5 & +17.5 \\
MFA \cite{zhang:2021} & 81.8 & 40.2 & 85.3 & - & - & - & 38.0 & 33.9 & 82.3 & - & 82.0 & 73.7 & 41.1 & 87.8 & - & 56.6 & - & 46.3 & 63.8 & \underline{62.5} & - & - & - \\
RED \cite{chao:2021} & 88.6 & 46.6 & 83.7 & 22.6 & 4.1 & 35.0 & 35.9 & 36.1 & 82.8 & - & 81.3 & 61.6 & 32.1 & 87.9 & - & 52.7 & - & 31.9 & 57.6 & 59.9 & 52.5 & 35.3 & +17.2 \\
ProDA \cite{zhang:2021b} & 87.8 & 45.7 & 84.6 & 37.1 & 0.6 & 44.0 & 54.6 & 37.0 & 88.1 & - & 84.4 & 74.2 & 24.3 & 88.2 & - & 51.1 & - & 40.5 & 45.6 & 62.0 & \underline{55.5} & \underline{34.9} & \textbf{+20.6} \\
\textbf{Co-T (ours)} & 78.1 & 36.9 & 84.0 & 9.3 & 0.2 & 47.4 & 49.2 & 19.3 & 89.0 & - & 89.6 & 77.9 & 52.3 & 91.5 & - & 60.3 & - & 47.1 & 64.7 & \textbf{64.6} & \textbf{56.0} & \textbf{35.4} & \textbf{+20.6}  \\
\midrule
\multicolumn{24}{c}{GTAV  (Source) $\rightarrow$ Cityscapes} \\
\midrule
AdSegNet\cite{tsai:2018} & 86.5 & 36.0 & 79.9 & 23.4 & 23.3 & 23.9 & 35.2 & 14.8 & 83.4 & 33.3 & 75.6 & 58.5 & 27.6 & 73.7 & 32.5 & 35.4 & 3.9 & 30.1 & 28.1 & - & 42.4 & 36.6 & +5.8 \\
IntraDA \cite{pan:2020} & 90.6 & 36.1 & 82.6 & 29.5 & 21.3 & 27.6 & 31.4 & 23.1 & 85.2 & 39.3 & 80.2 & 59.3 & 29.4 & 86.4 & 33.6 & 53.9 & 0.0 & 32.7 & 37.6 & - & 46.3 & 36.6 & +9.7 \\
CBST \cite{zou:2018} & 89.6 & 58.9 & 78.5 & 33.0 & 22.3 & 41.4 & 48.2 & 39.2 & 83.6 & 24.3 & 65.4 & 49.3 & 20.2 & 83.3 & 39.0 & 48.6 & 12.5 & 20.3 & 35.3 & - & 47.0 & 35.4 & +11.6 \\
CRST \cite{zou:2019} & 91.7 & 45.1 & 80.9 & 29.0 & 23.4 & 43.8 & 47.1 & 40.9 & 84.0 & 20.0 & 60.6 & 64.0 & 31.9 & 85.8 & 39.5 & 48.7 & 25.0 & 38.0 & 47.0 & - & 49.8 & 35.4 & +14.4  \\
DACS \cite{tranheden:2021} & 89.9 & 39.6 & 87.8 & 30.7 & 39.5 & 38.5 & 46.4 & 52.7 & 87.9 & 43.9 & 88.7 & 67.2 & 35.7 & 84.4 & 45.7 & 50.1 & 0.0 & 27.2 & 33.9 & - & 52.1 & 32.8 & +19.3 \\
DSP \cite{gao:2021} & 92.4 & 48.0 & 87.4 & 33.4 & 35.1 & 36.4 & 41.6 & 46.0 & 87.7 & 43.2 & 89.8 & 66.6 & 32.1 & 89.9 & 57.0 & 56.1 & 0.0 & 44.1 & 57.8 & - & 55.0 & 36.6 & +18.4 \\
MFA \cite{zhang:2021} & 94.5 & 61.1 & 87.6 & 41.4 & 35.4 & 41.2 & 47.1 & 45.7 & 86.6 & 36.6 & 87.0 & 70.1 & 38.3 & 87.2 & 39.5 & 54.7 & 0.3 & 45.4 & 57.7 & - & 55.7 & \textbf{45.6} & +10.1 \\
RED \cite{chao:2021} & 94.4 & 60.9 & 88.0 & 39.4 & 41.8 & 43.2 & 49.0 & 56.0 & 88.0 & 45.8 & 87.7 & 67.5 & 38.0 & 90.0 & 57.6 & 51.9 & 0.0 & 46.5 & 55.2 & - & \underline{57.9} & 34.8 & \underline{+23.1}\\
ProDA \cite{zhang:2021b} & 87.8 & 56.0 & 79.7 & 46.3 & 44.8 & 45.6 & 53.5 & 53.5 & 88.6 & 45.2 & 82.1 & 70.7 & 39.2 & 88.8 & 45.5 & 59.4 & 1.0 & 48.9 & 56.4 & - & 57.5 & \underline{36.6} & +20.9 \\
\textbf{Co-T (ours)} & 89.9 & 51.0 & 89.0 & 40.0 & 34.2 & 51.6 & 56.5 & 51.3 & 89.5 & 50.1 & 89.8 & 71.8 & 46.5 & 90.9 & 55.7 & 56.7 & 0.0 & 52.6 & 64.2 & - & \textbf{59.5} & 28.5 & \textbf{+31.0}  \\
\midrule
\multicolumn{24}{c}{Synscapes  (Source) $\rightarrow$ Cityscapes} \\
\midrule
AdSegNet\cite{tsai:2018} & 94.2 & 60.9 & 85.1 & 29.1 & 25.2 & 38.6 & 43.9 & 40.8 & 85.2 & 29.7 & 88.2 & 64.4 & 40.6 & 85.8 & 31.5 & 43.0 & 28.3 & 30.5 & 56.7 & - & 52.7 & \textbf{45.3} & +7.4 \\
IntraDA \cite{pan:2020} & 94.0 & 60.0 & 84.9 & 29.5 & 26.2 & 38.5 & 41.6 & 43.7 & 85.3 & 31.7 & 88.2 & 66.3 & 44.7 & 85.7 & 30.7 & 53.0 & 29.5 & 36.5 & 60.2 & - & \underline{54.2} & \textbf{45.3} & \underline{+8.9} \\
\textbf{Co-T (ours)} & 91.4 & 55.7 & 81.6 & 34.5 & 38.9 & 53.6 & 64.7 & 67.4 & 91.0 & 48.7 & 93.4 & 77.5 & 42.4 & 93.1 & 18.3 & 20.8 & 1.2 & 60.0 & 74.2 & - & \textbf{58.3} & \underline{45.0} & \textbf{+13.3} \\
\midrule
\multicolumn{24}{c}{GTAV + Synscapes  (Source) $\rightarrow$ Cityscapes} \\
\midrule
% X &  &  &  &  &  &  &  &  &  &  &  &  &  &  &  &  &  &  &  &  &  & \\
MADAN \cite{zhao:2019} & 94.1 & 61.0 & 86.4 & 43.3 & 32.1 & 40.6 & 49.0 & 44.4 & 87.3 & 47.7 & 89.4 & 61.7 & 36.3 & 87.5 & 35.5 & 45.8 & 31.0 & 33.5 & 52.1 & - & 55.7 & \textbf{51.6} & +4.1 \\
MsDACL \cite{he:2021} & 93.6 & 59.6 & 87.1 & 44.9 & 36.7 & 42.1 & 49.9 & 42.5 & 87.7 & 47.6 & 89.9 & 63.5 & 40.3 & 88.2 & 41.0 & 58.3 & 53.1 & 37.9 & 57.7 & - & \underline{59.0} & \textbf{51.6} & \underline{+7.4}\\
\textbf{Co-T (ours) }& 96.3 & 74.7 & 90.4 & 48.8 & 49.1 & 58.3 & 61.5 & 67.0 & 90.7 & 54.7 & 93.5 & 79.4 & 57.7 & 90.4 & 45.6 & 85.1 & 59.9 & 60.4 & 70.9 & - & \textbf{70.2} & \underline{50.0} & \textbf{+20.2} \\ 
\bottomrule
\end{tabular}}
\end{table*}

\subsection{Comparison with the state of the art}

In \Tab{sota_comp} we compare our co-training procedure with state-of-the-art methods when using Cityscapes as target domain. We divide the results into four blocks according to the source images we use: SYNTHA, GTAV, Synscapes, or GTAV+Synscapes. Most works in the literature present their results only using GTAV or SYNTHIA as source data. We obtain the best results in the SYNTHIA case, with $56$ mIoU (19 classes), and for GTAV with $59.5$ mIoU. On the other hand, each proposal from the literature uses their own CNN architecture and pre-trained models. Thus, we have added the mIoU score of the baseline that each work uses as starting point to improve according to the corresponding proposed method. Then, we show the difference between the final achieved mIoU score and the baseline one. In \Tab{sota_comp} this corresponds to column $\Delta$(Diff.). Note how our method reaches $20.6$ and $31.0$ points of mIoU increment on SYNTHIA and GTAV, respectively. The highest for GTAV, and the highest for SYNTHIA on pair with ProDA proposal. Additionally, for the shake of completeness, we have added the mIoU scores for the 13 classes setting of SYNTHIA since it is also a common practice in the literature. We can see that co-training obtains the best mIoU too. On the other hand, we are mostly interested in the 19 classes setting. Using Synscapes as source data we achieved state-of-the-art results in both $\Delta$(Diff.) ($13.3$ points) and final mIoU score ($58.3$). Note that, in this case, our baseline score is similar to the ones reported in previous literature. 

By performing a different LAB transform for each synthetic dataset individually, our co-training procedure allows to joint them as if they were one single domain. Thus, we have considered this setting too. Preliminary baseline experiments ({\ie}, without performing co-training) showed that the combinations GTAV + Synscapes and GTAV + Synscapes + SYNTHIA are the best performing, with a very scarce mIoU difference between them ($0.62$). Thus, for the shake of bounding the number of experiments, we have chosen GTAV + Synscapes as the only case combining datasets, so also avoiding the problem of the 19 {\vs} 16 classes discrepancy when SYNTHIA is combined with them. In fact, using GTAV + Synscapes, we reach a $\Delta$(Diff.) of $20.2$ points, with a final mIoU of $70.2$, which outperforms the second best in $11.2$ points, and it clearly improves the mIoU with respect to the use of these synthetic datasets separately ($15.6$ points comparing to GTAV, $11.9$ for Synscapes). Again, in this case, our baseline score is similar to the ones reported in previous literature.

\begin{table*}[t]
\caption{Co-training results compared to baseline (Source), LAB adjustment pre-processing (SrcLAB), self-training stage, and upper-bound (SrcLAB + Target). Note that Target refers to using the 100\% of the target domain images labeled for training by a human oracle. CB correspond to the class balance policy applied on GTAV. We remind also that, before running our co-training procedure (self-training stage and co-training loop), we apply target LAB adjustment on the synthetic datasets. In this table, the target domain is always Cityscapes.}
\label{tab:results_details}
\centering
\setlength{\tabcolsep}{0.8mm} 
\scalebox{0.84}[0.84]{\begin{tabular}{l||ccccccccccccccccccc||c||}
Methods & \rotatebox[origin=c]{90}{Road} & \rotatebox[origin=c]{90}{Sidewalk} & \rotatebox[origin=c]{90}{Building} & \rotatebox[origin=c]{90}{Wall} & \rotatebox[origin=c]{90}{Fence} & \rotatebox[origin=c]{90}{Pole} & \rotatebox[origin=c]{90}{Traffic light} & \rotatebox[origin=c]{90}{Traffic Sign} & \rotatebox[origin=c]{90}{Vegetation} & \rotatebox[origin=c]{90}{Terrain} & \rotatebox[origin=c]{90}{Sky} & \rotatebox[origin=c]{90}{Person} & \rotatebox[origin=c]{90}{Rider} & \rotatebox[origin=c]{90}{Car} & \rotatebox[origin=c]{90}{Truck} & \rotatebox[origin=c]{90}{Bus} & \rotatebox[origin=c]{90}{Train} & \rotatebox[origin=c]{90}{Motorbike} & \rotatebox[origin=c]{90}{Bike} & mIoU  \\
\midrule
\multicolumn{21}{c}{SYNTHIA (Source) $\rightarrow$ Cityscapes} \\
\midrule
Source & 38.47 & 17.42 & 75.39 & 4.92 & 0.22 & 30.58 & 17.79 & 15.48 & 73.86 & - & 80.55 & 63.28 & 22.57 & 62.47 & - & 27.12 & - & 13.22 & 23.92 & 35.46  \\
SrcLAB & 58.5 & 22.23 & 78.33 & 6.03 & 0.28 & 37.8 & 17.0 & 15.64 & 78.01 & - & 79.36 & 63.05 & 22.48 & 78.91 & - & 32.1 & - & 13.31 & 28.8 & 39.48  \\
Self-training stage (ours) & 72.14 & 30.37 & 82.97 & 2.97 & 0.11 & 43.72 & 34.49 & 14.00 & 87.09 & - & 86.87 & 73.82 & 43.45 & 87.57 & - & 42.44 & - & 21.69 & 56.36 &  48.74 \\
Co-training proce. (ours) & 78.14 & 36.98 & 84.07 & 9.34 & 0.28 & 47.49 & 49.2 & 19.35 & 89.07 & - & 89.62 & 77.92 & 52.32 & 91.50 & - & 60.37 & - & 47.10 & 64.76 & 56.09  \\
SrcLAB + Target & 97.92 & 84.42 & 92.60 & 53.87 & 61.70 & 65.93 & 70.67 & 78.00 & 92.71 & (65.68) & 94.98 & 83.29 & 66.49 & 95.32 & (77.37) & 88.20 & (71.84) & 67.45 & 78.13 & 79.48  \\
\midrule
$\Delta$(SrcLAB vs. Source) & 20.03 & 4.81 & 2.94 & 1.11 & 0.06 & 7.22 & -0.79 & 0.16 & 4.15 & - & -1.19 & -0.23 & -0.09 & 16.44 & - & 4.98 & - & 0.09 & 4.88 & 4.02  \\
$\Delta$(Co-t vs. Source) & 39.62 & 18.38 & 8.24 & 4.33 & 0.06 & 16.84 & 31.41 & 3.87 & 13.81 & - & 9.04 & 14.6 & 29.66 & 26.29 & - & 19.71 & - & 33.86 & 40.74 & 19.40 \\
$\Delta$(Co-t vs. SrcLAB) & 19.64 & 14.75 & 5.74 & 3.31 & 0 & 9.69 & 32.2 & 3.71 & 11.06 & - & 10.26 & 14.87 & 29.84 & 12.59 & - & 28.27 & - & 33.79 & 35.96 & 16.6 \\
$\Delta$(Co-t vs. Self-t) & 6.00 & 6.61 & 1.10 & 6.37 & 0.17 & 3.77 & 14.71 & 5.35 & 1.98 & - & 2.75 & 4.10 & 8.87 & 3.93 & - & 17.93 & - & 25.41 & 8.40 & 7.34  \\
$\Delta$(Co-t vs. SrcLAB + Tgt) & -19.83 & -48.62 & -8.97 & -44.62 & -61.42 & -18.51 & -21.47 & -58.65 & -5.04 & - & -5.39 & -5.41 & -14.26 & -6.56 & - & -41.37 & - & -20.37 & -13.47 & -24.62  \\
\midrule
\multicolumn{21}{c}{GTAV (Source) $\rightarrow$ Cityscapes} \\
\midrule
Source & 51.85 & 13.57 & 64.71 & 8.19 & 15.86 & 14.39 & 31.66 & 10.86 & 71.95 & 6.91 & 38.21 & 55.65 & 22.21 & 72.40 & 32.62 & 9.76 & 0.0 & 9.93 & 11.14 & 28.52  \\
SrcLAB & 75.25 & 23.38 & 76.59 & 19.72 & 16.86 & 32.28 & 28.37 & 13.73 & 81.75 & 25.47 & 46.71 & 64.0 & 31.76 & 84.14 & 32.29 & 16.23 & 0.08 & 23.41 & 27.27 & 37.86  \\
SrcLAB + CB & 73.34 & 26.30 & 73.50 & 29.57 & 21.16 & 35.04 & 42.78 & 20.09 & 84.64 & 26.48 & 53.20 & 63.02 & 40.77 & 81.90 & 34.16 & 31.56 & 4.74 & 36.05 & 34.07 & 42.76  \\
Self-training stage (ours) & 85.31 & 36.82 & 85.11 & 41.09 & 25.62 & 46.39 & 45.19 & 33.44 & 88.98 & 45.55 & 72.99 & 69.54 & 42.43 & 89.36 & 44.42 & 57.5 & 1.28 & 45.51 & 59.78 & 53.49 \\
Co-training proce. (ours) & 89.92 & 51.03 & 89.09 & 40.05 & 34.23 & 51.61 & 56.54 & 51.36 & 89.50 & 50.12 & 89.83 & 71.88 & 46.50 & 90.91 & 55.72 & 56.77 & 0.0 & 52.61 & 64.21 & 59,57   \\
SrcLAB + Target & 98.20 & 85.43 & 92.74 & 59.07 & 63.05 & 65.26 & 69.43 & 77.10 & 92.63 & 65.26 & 94.70 & 82.11 & 63.22 & 95.22 & 85.05 & 86.07 & 67.27 & 64.84 & 77.21 & 78.10  \\
\midrule
$\Delta$(SrcLAB + CB vs. Source) & 21.49 & 12.73 & 8.79 & 21.38 & 5.3 & 20.65 & 11.12 & 9.23 & 12.69 & 19.57 & 14.99 & 7.37 & 18.56 & 9.5 & 1.54 & 21.8 & 4.74 & 26.12 & 22.93 & 14.24  \\
$\Delta$(Co-t vs. Source) & 38.07 & 37.46 & 24.38 & 31.86 & 18.37 & 37.22 & 24.88 & 40.5 & 17.55 & 43.21 & 51.62 & 16.23 & 24.29 & 18.51 & 23.1 & 47.01 & 0.0 & 42.68 & 53.07 & 31.05  \\
$\Delta$(Co-t vs. SrcLAB + Class Balance) & 16.58 & 24.73 & 15.59 & 10.48 & 13.07 & 16.57 & 13.76 & 31.27 & 4.86 & 23.64 & 36.63 & 8.86 & 5.73 & 9.01 & 21.56 & 25.21 & -4.74 & 16.56 & 30.14 & 16.81  \\
$\Delta$(Co-t vs. Self-t) & 4.61 & 14.21 & 3.98 & -1.04 & 8.61 & 5.22 & 11.35 & 17.92 & 0.52 & 4.57 & 16.84 & 2.34 & 4.07 & 1.55 & 11.3 & -0.73 & -1.28 & 7.1 & 4.43 & 6.08   \\
$\Delta$(Co-t vs. SrcLAB + Tgt) & -8.28 & -34.4 & -3.65 & -19.02 & -28.82 & -13.65 & -12.89 & -25.74 & -3.13 & -15.14 & -4.87 & -10.23 & -16.72 & -4.31 & -29.33 & -29.3 & -67.27 & -12.23 & -13.0 & -18.53  \\
\midrule
\multicolumn{21}{c}{Synscapes (Source) $\rightarrow$ Cityscapes} \\
\midrule
Source & 83.81 & 42.15 & 61.87 & 26.10 & 21.69 & 44.65 & 47.12 & 53.86 & 81.30 & 33.57 & 53.53 & 67.79 & 29.68 & 85.66 & 14.81 & 6.66 & 2.36 & 34.94 & 63.53 & 45.01  \\
SrcLAB & 78.39 & 37.47 & 67.39 & 16.45 & 19.09 & 48.5 & 51.79 & 58.54 & 83.18 & 29.89 & 64.79 & 70.17 & 29.27 & 85.39 & 18.42 & 10.42 & 3.32 & 36.48 & 64.61 & 45.98  \\

Self-training stage (ours) & 89.55 & 50.19 & 84.26 & 33.61 & 37.67 & 57.29 & 60.11 & 64.00 & 90.61 & 47.13 & 91.22 & 72.15 & 21.17 & 91.99 & 15.38 & 20.09 & 9.35 & 44.94 & 70.78 & 55.34  \\
Co-training proce. (ours) & 91.46 & 55.76 & 81.63 & 34.58 & 38.92 & 53.66 & 64.74 & 67.43 & 91.02 & 48.72 & 93.45 & 77.54 & 42.40 & 93.14 & 18.35 & 20.84 & 1.29 & 60.03 & 74.22 & 58.38  \\
SrcLAB + Target & 98.03 & 84.49 & 92.90 & 59.10 & 63.70 & 67.18 & 71.67 & 79.50 & 92.74 & 65.51 & 94.81 & 83.93 & 68.07 & 95.45 & 82.89 & 91.83 & 83.79 & 70.91 & 79.24 & 80.30  \\
\midrule
$\Delta$(SrcLAB vs. Source) & 0.97 & -5.42 & -4.68 & 5.52 & -9.65 & -2.6 & 3.85 & 4.67 & 4.68 & 1.88 & -3.68 & 11.26 & 2.38 & -0.41 & -0.27 & 3.61 & 3.76 & 0.96 & 1.54 & 1.08  \\
$\Delta$(Co-t vs. Source) & 7.65 & 13.61 & 19.76 & 8.48 & 17.23 & 9.01 & 17.62 & 13.57 & 9.72 & 15.15 & 39.92 & 9.75 & 12.72 & 7.48 & 3.54 & 14.18 & -1.07 & 25.09 & 10.69 & 13.37 \\
$\Delta$(Co-t vs. SrcLAB) & 13.07 & 18.29 & 14.24 & 18.13 & 19.83 & 5.16 & 12.95 & 8.89 & 7.84 & 18.83 & 28.66 & 7.37 & 13.13 & 7.75 & -0.07 & 10.42 & -2.03 & 23.55 & 9.61 & 12.40  \\
$\Delta$(Co-t vs. Self-t) & 1.91 & 5.57 & -2.63 & 0.97 & 1.25 & -3.63 & 4.63 & 3.43 & 0.41 & 1.59 & 2.23 & 5.39 & 21.23 & 1.15 & 2.97 & 0.75 & -8.06 & 15.09 & 3.44 &  3.04 \\
$\Delta$(Co-t vs. SrcLAB + Tgt) & -6.57 & -28.73 & -11.27 & -24.52 & -24.78 & -13.52 & -6.93 & -12.07 & -1.72 & -16.79 & -1.36 & -6.39 & -25.67 & -2.31 & -64.54 & -70.99 & -82.5 & -10.88 & -5.02 & -21.92 \\
\midrule
\multicolumn{21}{c}{GTAV + Synscapes (Source) $\rightarrow$ Cityscapes} \\
\midrule
% X &  &  &  &  &  &  &  &  &  &  &  &  &  &  &  &  &  &  &  &   \\
Source & 66.39 & 33.54 & 79.58 & 29.43 & 40.24 & 49.73 & 56.12 & 46.51 & 81.22 & 18.40 & 79.06 & 73.18 & 29.67 & 85.25 & 43.00 & 6.46 & 23.02 & 47.71 & 61.63 & 50.01  \\
SrcLAB & 87.97 & 47.45 & 85.14 & 34.31 & 43.16 & 49.82 & 57.16 & 47.85 & 88.88 & 45.00 & 82.53 & 72.58 & 38.22 & 89.16 & 51.91 & 61.31 & 40.06 & 43.64 & 60.85 & 59.32  \\ 
Self-training stage (ours) & 93.93 & 66.08 & 89.95 & 46.40 & 48.13 & 56.30 & 59.65 & 65.16 & 90.25 & 52.22 & 93.33 & 75.97 & 41.15 & 90.40 & 44.98 & 75.08 & 65.52 & 55.52 & 71.98 & 67.47  \\ 
Co-training proce. (ours) & 96.30 & 74.72 & 90.44 & 48.89 & 49.15 & 58.36 & 61.52 & 67.05 & 90.75 & 54.75 & 93.52 & 79.48 & 57.71 & 90.48 & 45.61 & 85.11 & 59.95 & 60.41 & 70.96 & 70.23 \\ 
SrcLAB + Target & 97.88 & 83.43 & 92.33 & 64.15 & 61.77 & 63.45 & 68.04 & 75.17 & 92.31 & 62.74 & 94.08 & 82.00 & 64.16 & 95.01 & 84.25 & 89.66 & 75.56 & 63.28 & 76.07 & 78.18 \\
\midrule
$\Delta$(SrcLAB vs. Source) & 21.58 & 13.91 & 5.56 & 4.88 & 2.92 & 0.09 & 1.04 & 1.34 & 7.66 & 26.6 & 3.47 & -0.6 & 8.55 & 3.91 & 8.91 & 54.85 & 17.04 & -4.07 & -0.78 & 9.31  \\
$\Delta$(Co-t vs. Source) & 29.91 & 41.18 & 10.86 & 19.46 & 8.91 & 8.63 & 5.40 & 20.54 & 9.53 & 36.35 & 14.46 & 6.3 & 28.04 & 5.23 & 2.61 & 78.65 & 36.93 & 12.70 & 9.33 & 20.22 \\
$\Delta$(Co-t vs. SrcLAB) & 8.33 & 27.27 & 5.30 & 14.58 & 5.99 & 8.54 & 4.36 & 19.20 & 1.87 & 9.75 & 10.99 & 6.9 & 19.49 & 1.32 & -6.3 & 23.8 & 19.89 & 16.77 & 9.28 & 10.91  \\
$\Delta$(Co-t vs. Self-t) & 2.37 & 8.64 & 0.49 & 2.49 & 1.02 & 2.06 & 1.87 & 1.89 & 0.5 & 2.53 & 0.19 & 3.51 & 16.56 & 0.08 & 0.63 & 10.03 & -5.57 & 4.89 & -1.85 & 2.76 \\
$\Delta$(Co-t vs. SrcLAB + Tgt) & -1.58 & -8.71 & -1.89 & -15.26 & -12.62 & -5.09 & -6.52 & -8.12 & -1.56 & -7.99 & -0.56 & -2.52 & -6.45 & -4.53 & -38.64 & -4.55 & -15.61 & -2.87 & -5.94 & -7.95 \\
\bottomrule
\end{tabular}}
\end{table*}

\begin{table}[t]
\caption{Contribution of the main components of our proposal. Case study: GTAV + Synscapes $\rightarrow$ Cityscapes.}
\centering
\setlength{\tabcolsep}{1.0mm}
\scalebox{0.85}[0.85]{
\begin{tabular}{@{}lcccccc@{}} \toprule
     &          &       & \multicolumn{3}{c}{Co-training Procedure} \\ \cmidrule(r){4-6}
     &          &       & \multicolumn{2}{c}{Self-training Stage}   \\ \cmidrule(r){4-5}
     & Baseline & + LAB & + MixBatch & + ClassMix & + Co-training loop & Upper-bound \\ \midrule
mIoU & 50.01    & 59.32 & 66.18      & 67.47      & 70.23              & 78.18 \\
Gain & -        & +9.31 & +6.86      & +1.29      & +2.76              & - \\ \bottomrule
\end{tabular}}
\label{tab:ablation}
\end{table}

\begin{table*}[t]
\caption{Analogous to \Tab{results_details} with BDD100K and Mapillary as target domains.}
\label{tab:other_real}
\centering
\setlength{\tabcolsep}{1.0mm} 
\scalebox{0.865}[0.865]{\begin{tabular}{l||ccccccccccccccccccc||c||}
Methods & \rotatebox[origin=c]{90}{Road} & \rotatebox[origin=c]{90}{Sidewalk} & \rotatebox[origin=c]{90}{Building} & \rotatebox[origin=c]{90}{Wall} & \rotatebox[origin=c]{90}{Fence} & \rotatebox[origin=c]{90}{Pole} & \rotatebox[origin=c]{90}{Traffic light} & \rotatebox[origin=c]{90}{Traffic Sign} & \rotatebox[origin=c]{90}{Vegetation} & \rotatebox[origin=c]{90}{Terrain} & \rotatebox[origin=c]{90}{Sky} & \rotatebox[origin=c]{90}{Person} & \rotatebox[origin=c]{90}{Rider} & \rotatebox[origin=c]{90}{Car} & \rotatebox[origin=c]{90}{Truck} & \rotatebox[origin=c]{90}{Bus} & \rotatebox[origin=c]{90}{Train} & \rotatebox[origin=c]{90}{Motorbike} & \rotatebox[origin=c]{90}{Bike} & mIoU  \\
\midrule
\multicolumn{21}{c}{GTAV + Synscapes (Source) $\rightarrow$ BDD100K} \\
\midrule
Source & 67.83 & 20.84 & 54.86 & 9.00 & 27.57 & 30.24 & 31.74 & 20.75 & 62.69 & 15.39 & 63.75 & 54.53 & 24.08 & 65.92 & 12.82 & 9.10 & 0.07 & 39.58 & 39.04 & 34.20 \\
SrcLAB & 74.22 & 26.07 & 68.48 & 7.94 & 15.51 & 31.09 & 38.69 & 22.90 & 69.33 & 25.92 & 74.27 & 59.35 & 18.81 & 72.79 & 23.66 & 19.75 & 0.02 & 54.72 & 35.48 & 38.68 \\
Self-training stage (ours) & 88.52 & 26.21 & 78.77 & 14.48 & 35.41 & 41.40 & 49.27 & 31.74 & 75.86 & 35.89 & 88.85 & 60.39 & 35.22 & 85.48 & 35.04 & 42.29 & 0.00 & 51.28 & 47.40 & 48.60 \\
Co-training proce. (ours) & 88.43 & 31.63 & 80.05 & 13.05 & 39.89 & 41.81 & 46.12 & 29.67 & 76.06 & 37.79 & 89.50 & 63.08 & 39.94 & 85.78 & 40.52 & 42.71 & 0.07 & 53.60 & 50.40 & 50.11  \\
SrcLAB + Target & 93.33 & 60.89 & 84.41 & 31.45 & 47.74 & 49.62 & 55.33 & 47.77 & 85.00 & 42.77 & 92.60 & 66.10 & 38.91 & 88.11 & 40.63 & 71.09 & 0.00 & 57.71 & 54.90 & 58.33 \\
\midrule
$\Delta$(SrcLAB vs. Source) & 6.39 & 5.23 & 9.62 & -1.06 & -12.06 & 0.85 & 6.95 & 2.15 & 6.64 & 10.53 & 10.52 & 4.82 & -5.27 & 6.87 & 10.84 & 10.65 & -0.05 & 15.14 & -3.56 & 4.48  \\
$\Delta$(Co-t vs. Source) & 25.5 & 40.05 & 29.55 & 22.45 & 20.17 & 19.38 & 23.59 & 27.02 & 22.31 & 27.38 & 28.85 & 11.57 & 14.83 & 22.19 & 27.81 & 61.99 & -0.07 & 18.13 & 15.86 & 24.13 \\
$\Delta$(Co-t vs. SrcLAB) & 19.11 & 34.82 & 19.93 & 23.51 & 32.23 & 18.53 & 16.64 & 24.87 & 15.67 & 16.85 & 18.33 & 6.75 & 20.1 & 15.32 & 16.97 & 51.34 & -0.02 & 2.99 & 19.42 & 19.65  \\
$\Delta$(Co-t vs. Self-t) & -0.09 & 5.42 & 1.28 & -1.43 & 4.48 & 0.41 & -3.15 & -2.07 & 0.2 & 1.9 & 0.65 & 2.69 & 4.72 & 0.3 & 5.48 & 0.42 & 0.07 & 2.32 & 3.0 &  1.51 \\
$\Delta$(Co-t vs. SrcLAB + Tgt) & -4.9 & -29.26 & -4.36 & -18.4 & -7.85 & -7.81 & -9.21 & -18.1 & -8.94 & -4.98 & -3.1 & -3.02 & 1.03 & -2.33 & -0.11 & -28.38 & 0.07 & -4.11 & -4.5 & -8.22  \\
\midrule
\multicolumn{21}{c}{GTAV + Synscapes (Source) $\rightarrow$ Mapillary Vistas} \\
\midrule
% X &  &  &  &  &  &  &  &  &  &  &  &  &  &  &  &  &  &  &  &   \\
Source & 68.81 & 31.73 & 68.88 & 25.20 & 37.94 & 38.79 & 49.79 & 20.57 & 73.27 & 29.66 & 80.62 & 63.81 & 42.75 & 80.65 & 35.74 & 16.86 & 1.85 & 44.56 & 47.09 & 45.19 \\
SrcLAB & 72.62 & 43.18 & 70.89 & 17.21 & 25.18 & 35.05 & 57.74 & 55.73 & 76.78 & 27.09 & 88.72 & 71.34 & 24.34 & 77.89 & 46.29 & 47.37 & 0.00 & 34.27 & 46.77 & 48.34 \\
Self-training stage (ours) & 89.44 & 53.30 & 85.28 & 36.57 & 44.89 & 47.10 & 59.18 & 65.94 & 84.58 & 48.25 & 97.44 & 74.23 & 55.71 & 89.37 & 58.34 & 59.45 & 1.47 & 49.44 & 51.50 & 60.60 \\
Co-training proce. (ours) & 90.44 & 57.83 & 85.59 & 36.38 & 45.56 & 49.64 & 59.73 & 67.62 & 84.27 & 47.08 & 96.79 & 74.80 & 56.05 & 90.42 & 56.34 & 49.86 & 10.71 & 49.62 & 55.94 & 61.30 \\
SrcLAB + Target & 94.02 & 69.46 & 88.70 & 51.38 & 60.17 & 57.59 & 64.21 & 75.16 & 90.70 & 69.35 & 98.27 & 76.02 & 56.70 & 91.42 & 60.49 & 73.35 & 33.81 & 60.87 & 66.63 & 70.44 \\
\midrule
$\Delta$(SrcLAB vs. Source) & -2.93 & -0.24 & 0.73 & -8.10 & -8.64 & 5.25 & 5.94 & 21.32 & 3.34 & 2.73 & 1.74 & 3.47 & 4.75 & 1.86 & 5.84 & -3.15 & 1.09 & 7.51 & 6.69 & 2.59  \\
$\Delta$(Co-t vs. Source) & 21.63 & 26.10 & 16.71 & 11.18 & 7.62 & 10.85 & 9.94 & 47.05 & 11.0 & 17.42 & 16.17 & 10.99 & 13.30 & 9.77 & 20.60 & 33.0 & 8.86 & 5.06 & 8.85 & 16.11 \\
$\Delta$(Co-t vs. SrcLAB) & 24.56 & 26.34 & 15.98 & 19.28 & 16.26 & 5.6 & 4.0 & 25.73 & 7.66 & 14.69 & 14.43 & 7.52 & 8.55 & 7.91 & 14.76 & 36.15 & 7.77 & -2.45 & 2.16 &  13.52 \\
$\Delta$(Co-t vs. Self-t) & 1.0 & 4.53 & 0.31 & -0.19 & 0.67 & 2.54 & 0.55 & 1.68 & -0.31 & -1.17 & -0.65 & 0.57 & 0.34 & 1.05 & -2.0 & -9.59 & 9.24 & 0.18 & 4.44 & 0.7  \\
$\Delta$(Co-t vs. SrcLAB + Tgt) & -3.58 & -11.63 & -3.11 & -15.0 & -14.61 & -7.95 & -4.48 & -7.54 & -6.43 & -22.27 & -1.48 & -1.22 & -0.65 & -1.0 & -4.15 & -23.49 & -23.1 & -11.25 & -10.69 & -9.14 \\
\bottomrule
\end{tabular}}
\end{table*}

\subsection{Ablative study and Qualitative results}

In \Tab{results_details} we compare co-training results with corresponding baselines and upper-bounds. We also report the results of applying LAB adjustment as only UDA step, as well as the results from one of the models obtained after our self-training stage (we chose the model from the last cycle). Overall, in all cases the co-training loop (which completes the co-training procedure) improves the self-training stage, and this stage, in turn, improves over LAB adjustment. Moreover, when combining GTAV + Synscapes we are only $7.95$ mIoU points below the upper-bound, after improving $\sim20.22$ mIoU points the baseline. 

To complement our experimental analysis, we summarize in \Tab{ablation} the contribution of the main components of our proposal for the case GTAV + Synscapes $\rightarrow$ Cityscapes. First, we can see how a proper pre-processing of the data is relevant. In particular, performing synth-to-real LAB space alignment already allows to improve $9.31$ points of mIoU. This contribution can also be seen in Tables \ref{tab:results_details} and \ref{tab:other_real}, where improvements range from $2.59$ mIoU points (GTAV+Synscapes$\rightarrow$Mapillary Vistas) to $9.34$ (GTAV$\rightarrow$Cityscapes). This LAB adjustment is a step hardly seen in synth-to-real UDA literature which should not be ignored. Then, back to \Tab{ablation}, we see that properly combining labeled source images and pseudo-labeled target images (MixBatch) is also relevant since it provides an additional gain of $6.86$ points. Note that this MixBatch is basically the \emph{cool world} idea which we can trace back to work of our own lab done before the deep learning era in computer vision \cite{Vazquez:2011}. In addition, performing our ClassMix-inspired collage also contributes with $1.29$ points of mIoU, and the final collaboration of models returns $2.76$ additional points of mIoU. Overall, the main components of our synth-to-real UDA procedure contribute with $10.91$ points of mIoU and LAB alignment $9.31$ points. We conclude that all the components of the proposed procedure are relevant.

In order to confirm these positive results, we applied our method to two additional target domains which are relatively challenging, namely, Mapillary Vistas and BDD100K. In fact, up to the best of our knowledge, in the current literature there are not synth-to-real UDA semantic segmentation results reported for them. Our results can be seen in \Tab{other_real}, directly focusing on the combination of GTAV + Synscapes as source domain. In this case, the co-training loop improves less over the intermediate self-training stage. Still, for BDD100K the final mIoU is only $8.22$ mIoU points below the upper-bound, after improving $24.13$ mIoU points the baseline. For Mapillary Vistas our methods remains only $9.14$ mIoU points below the upper-bound and improves $16.11$ mIoU points the baseline. Up to the best of our knowledge, this are state-of-the-art results for BDD100K and Mapillary Vistas when addressing synth-to-real UDA semantic segmentation. 

\begin{figure*}[t!]
%[trim=left bottom right top, clip]
\centering
     \includegraphics[width=1.0\textwidth]{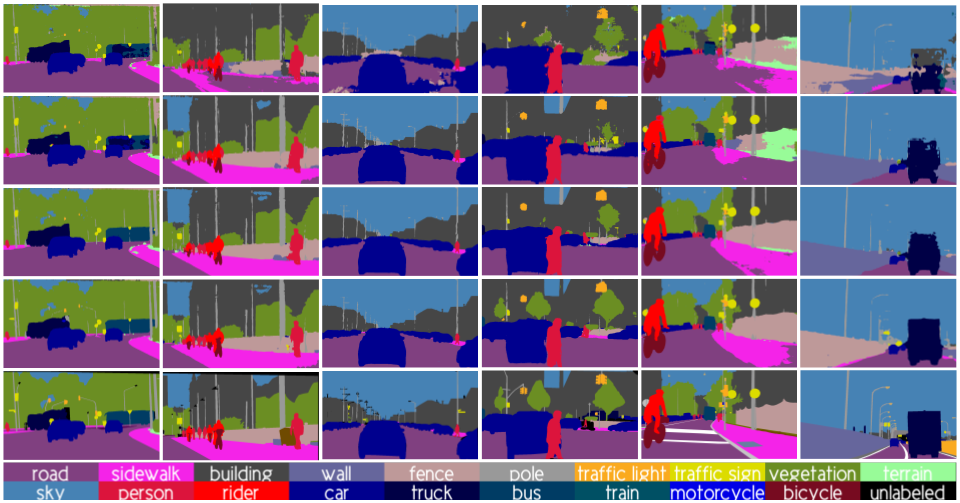}
\caption{Qualitative results using GTAV + Synscapes as source domain. From left to right, the two first columns correspond to Cityscapes in the role of target domain, next two  columns to BDD100K, and last two to Mapillary Vistas. Top to bottom rows correspond to SrcLAB, self-training stage, full co-training procedure, upper-bound, and ground truth, respectively.}
\label{fig:qualitative}
\end{figure*}

\begin{figure*}[h!]
%[trim=left bottom right top, clip]
\centering
     \includegraphics[width=1.0\textwidth]{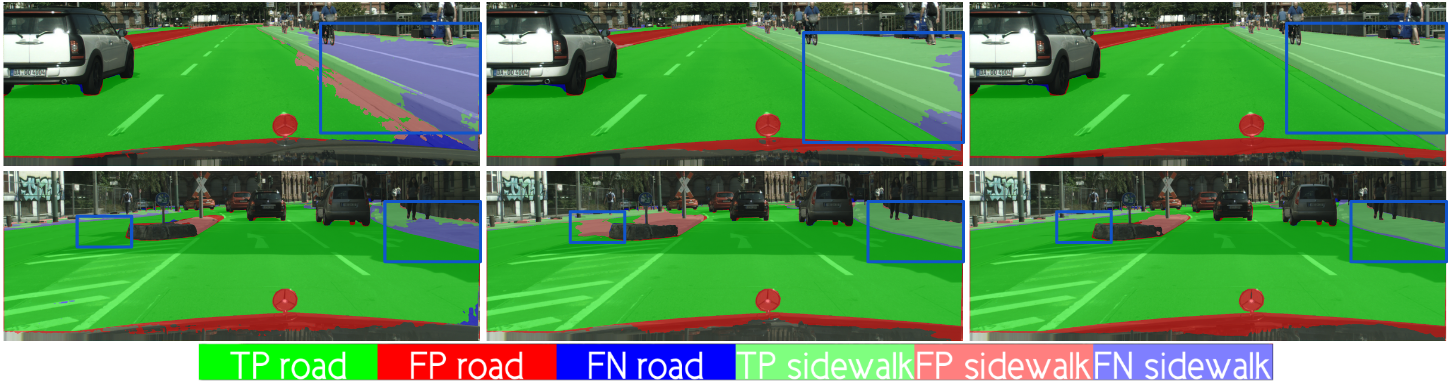}
\caption{Qualitative results (GTAV + Synscapes $\rightarrow$ Cityscapes) focusing on TP/FP/FN for road and sidewalk classes. Columns, left to right: SrcLAB, self-training stage, co-training loop (full co-training procedure). Blue boxes highlight areas of interest.}
\label{fig:qualitative_road_sidewalk}
\end{figure*}

\begin{figure*}[h!]
%[trim=left bottom right top, clip]
\centering
     \includegraphics[width=1.0\textwidth]{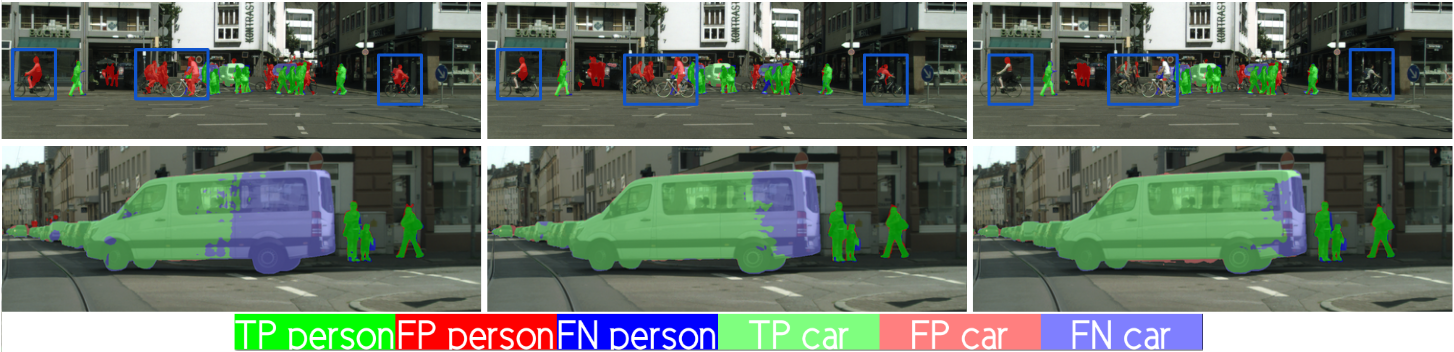}
\caption{Analogous to \Fig{qualitative_road_sidewalk} for the classes Person and Car.}
\label{fig:qualitative_person_car}
\end{figure*}

\Fig{qualitative} presents qualitative results of semantic segmentation for the different real-world (target) datasets, when using GTAV + Synscapes as source domain. We observe how the baselines have problems with dynamic objects ({\eg}, cars, trucks) and some infrastructure classes like sidewalk are noisy. The self-training stage mitigates the problems observed in the only-source (with LAB adjustment) results to a large extent. However, we can still observe instabilities in classes like truck or bus, which the co-training loop (full co-training procedure) achieves to address properly. Nevertheless, the co-training procedure is not perfect and several errors are observed in some classes preventing to reach upper-bound mIoU. In fact, upper-bounds are neither perfect, which is due to the difficulty of performing semantic segmentation in on-board images. 

Figures \ref{fig:qualitative_road_sidewalk} and \ref{fig:qualitative_person_car} exemplify these comments by showing the pseudo-labeling evolution for several classes of special interests such as Road, Sidewalk, Pedestrian, and Car. In \Fig{qualitative_road_sidewalk}, we see how the SrcLAB model has particular problems to segment well the sidewalk, however, the self-training stage resolves most errors although it may introduce new ones (mid-bottom image), while the co-training loop is able to recover from such errors. In \Fig{qualitative_person_car}, we can see (bottom row) how the self-training stage improves the pseudo-labeling of a van, while the co-training loop improves it even more. Analogously, we can see (top row) how self-training helps to alleviate the confusion between pedestrian and riders, while the co-training loop almost removes all the confusion errors between these two classes. 

\section{Conclusions}
\label{sec:conclusions}

In this paper, we have addressed the training of semantic segmentation models under the challenging setting of synth-to-real unsupervised domain adaptation (UDA), {\ie}, assuming access to a set of synthetic images (source) with automatically generated ground truth together with a set of unlabeled real-world images (target). We have proposed a new co-training procedure combining a self-training stage and a co-training loop where two models arising from the self-training stage collaborate for mutual improvement. The overall procedure treats the deep models as black boxes and drives their collaboration at the level of pseudo-labeled target images, {\ie}, neither modifying loss functions is required, nor explicit feature alignment. We have tested our proposal on standard synthetic (GTAV, Synscapes, SYNTHIA) and real-world datasets (Cityscapes, BDD100K, Mapillary Vistas). Our co-training shows improvements ranging from {$\sim$13} to {$\sim$31} mIoU points over baselines, remaining very closely (less than 10 points) to the upper-bounds. In fact, up to the best of our knowledge, we are the first reporting such kind of results for challenging target domains such as BDD100K and Mapillary Vistas. Moreover, we have shown how the different components of our co-training procedure contribute to improve final mIoU. Future work, will explore collaboration from additional perception models at the co-training loop, {\ie}, not necessarily based on semantic segmentation but such collaborations may arise from object detection or monocular depth estimation.\\
%NOTE: BDD100k have pick-up cars as trucks, some mistakes in the annotations (most in train) and some inconsistencies with borderline classes between consecutive frames   

\noindent\emph{\textbf{Comments:}} 
\begin{enumerate}
\item The code is publicly available at \url{https://github.com/JoseLGomez/Co-training_SemSeg_UDA}.
\item The paper was accepted on Sensors 2023, Special Issue Machine Learning for Autonomous Driving Perception and Prediction. Available at \url{https://www.mdpi.com/1424-8220/23/2/621}
\end{enumerate}
%{\appendix}

%\begin{thebibliography}{1}
\bibliographystyle{IEEEtran}

\bibliography{references}

%\end{thebibliography}


%\vspace{-1cm}
\begin{IEEEbiography}[{\includegraphics[height=1in,keepaspectratio]{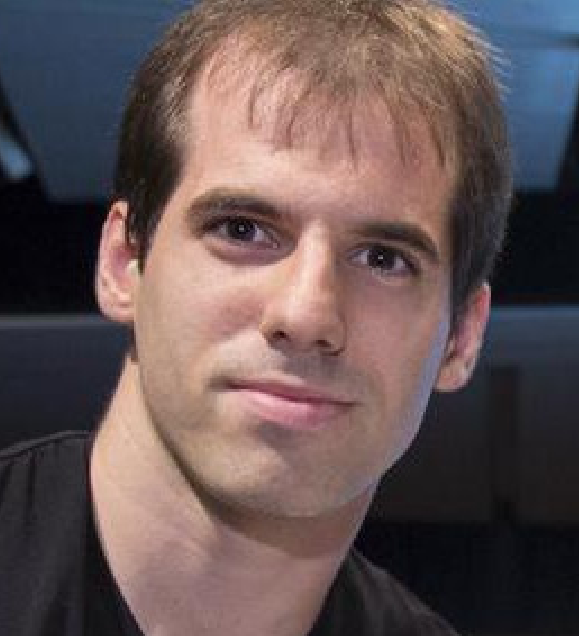}}]{Jose Luis G\'omez Zurita}
Received his B.S. degree in computer engineering and M.S. degree in computer vision from Universitat Autònoma de Barcelona (UAB) in 2016 and 2017 respectively. He is pursuing his Ph.D. in Computer Vision Center, Universitat Aut\'onoma de Barcelona. His research fields are autonomous driving, deep learning, and domain adaptation.
\end{IEEEbiography}

%\vspace{-1cm}
\begin{IEEEbiography}[{\includegraphics[height=1.25in,keepaspectratio]{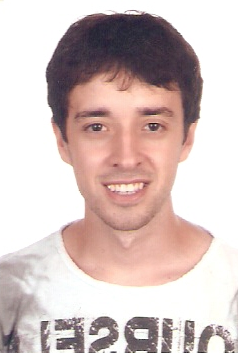}}]{Gabriel Villalonga}
%Received the B.Sc. degree in Computer Science from Universitat Autònoma de Barcelona (UAB) in 2014 where he performed his final project on 3D pedestrian detection for autonomous vehicles, supervised by A.M. Lopez and D. Vázquez from the Computer Vision Center (CVC). He received his M.Sc. on Computer Vision  at the UAB, UPC, UPF and UOC in 2015 doing his M.Sc. Thesis on the study of the pedestrian behaviour analysis, supervised by A.M. Lopez, D. Vázquez and G. Ros. Currently. He finished the Ph.D. on Informatics for his Thesis Leveraging synthetic data to create autonomous driving perception systems supervised by A.M. Lopez and G. Ros. He is an active member of the ADAS group and the perception lead in the TDA project at the Computer Vision Center, to develop an Autonomous Vehicle for rural environments.
Received the B.Sc. degree in Computer Science in 2014 and the M.Sc. degree on Computer Vision in 2015, both at Universitat Autònoma de Barcelona (UAB). Gabriel obtained the PhD degree at UAB for his dissertation \emph{Leveraging synthetic data to create autonomous driving perception systems}. He is member of the Autonomous Driving lab at the Computer Vision Center (CVC)/UAB. He focuses on developing autonomous driving for underpopulated rural environments.
\end{IEEEbiography}

%\vspace{-1cm}
\begin{IEEEbiography}
[{\includegraphics[height=3cm,keepaspectratio]{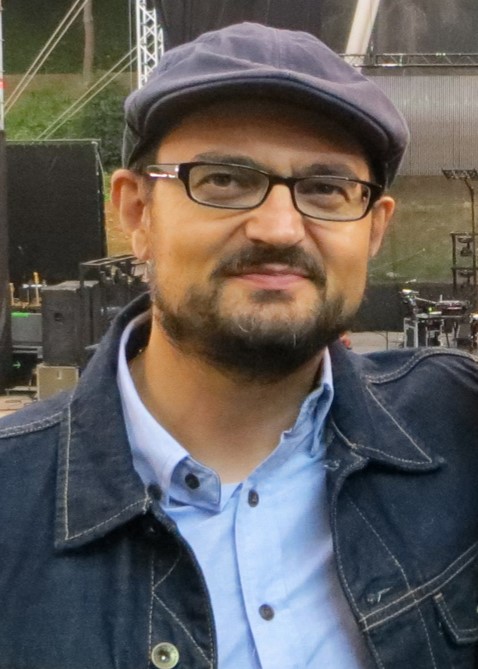}}]{\\ Antonio M. L\'opez} is the PI of the Autonomous Driving lab of the Computer Vision Center (CVC) at the Univ. Aut\`onoma de Barcelona (UAB). He carries research at the intersection of computer vision, machine learning, simulation, and autonomous driving. He is one of the fathers of the SYNTHIA dataset and the CARLA simulator. He collaborates with industry partners to bring state-of-the-art techniques to the field of autonomous driving. Antonio is granted by the Catalan ICREA Academia program.
\end{IEEEbiography}

\vfill

\end{document}